\documentclass[10pt,twocolumn,letterpaper]{article}

\usepackage[pagenumbers]{iccv} 

\definecolor{iccvblue}{rgb}{0.21,0.49,0.74}
\usepackage[pagebackref,breaklinks,colorlinks,allcolors=iccvblue]{hyperref}
\def\ModelName{HERMES}
\def\ModelNameExtI{temporal-co\textbf{HER}ent long-for\textbf{M} understanding with \textbf{E}pisodes and \textbf{S}emantics}
\def\SubmoduleI{ECO}
\def\SubmoduleIExt{Episodic COmpressor}
\def\SubmoduleII{SeTR}
\def\SubmoduleIIExt{Semantics reTRiever}

\usepackage{microtype}
\usepackage{graphicx}
\usepackage{subcaption}
\usepackage{booktabs} 
\usepackage{algorithm}    
\usepackage{algpseudocode}  
\usepackage{amsmath}      

\usepackage{amssymb}
\usepackage{multirow}

\def\paperID{13355} 
\def\confName{ICCV}
\def\confYear{2025}

\title{\ModelName: temporal-coHERent long-forM understanding with Episodes and Semantics}

\author{
    Gueter Josmy Faure\textsuperscript{1} \hspace{0.5em}
    Jia-Fong Yeh\textsuperscript{1} \hspace{0.5em}
    Min-Hung Chen\textsuperscript{2} \hspace{0.5em}
    Hung-Ting Su\textsuperscript{1} \\
    Shang-Hong Lai\textsuperscript{4} \hspace{0.5em}
    Winston H. Hsu\textsuperscript{1} \\ \\
    \textsuperscript{1}National Taiwan University \hspace{0.5em}
    \textsuperscript{2}NVIDIA \hspace{0.5em}
    \textsuperscript{4}National Tsing Hua University 
}

\begin{document}
\twocolumn[{%
	\renewcommand\twocolumn[1][]{#1}%
	\maketitle
	\begin{center}
		\newcommand{\teaserwidth}{\textwidth}
		\centerline{
			\includegraphics[width=\teaserwidth,clip]{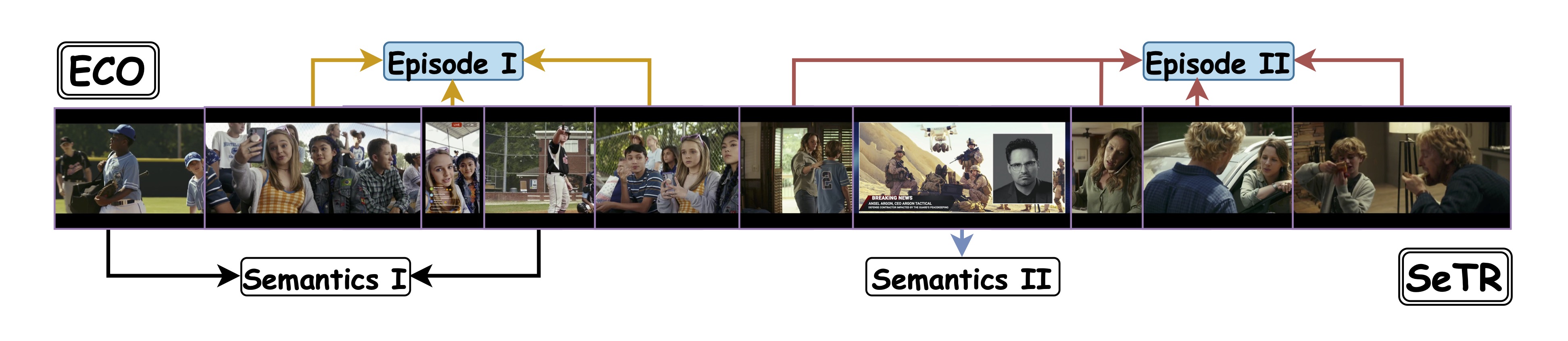}
		}
		\captionof{figure}{\textbf{Semantic Knowledge and Episodic Memory Aggregation:} Our \SubmoduleIExt\ (\SubmoduleI) processes and aggregates temporal information across different scales: (I) Social interactions among adolescents in an outdoor setting, and (II) Complex family dynamics portrayed through parent-child interactions. Simultaneously, our \SubmoduleIIExt\ (\SubmoduleII) extracts high-level semantic information: (I) The contextual environment of a baseball game, and (II) The intersection of media consumption and domestic life through a news broadcast. This dual-level approach enables comprehensive video understanding by capturing both specific events and overarching concepts.}
		\vspace{-0.05in}
		\label{fig:teaser}
	\end{center}%
}]
\begin{abstract}
Long-form video understanding presents unique challenges that extend beyond traditional short-video analysis approaches, particularly in capturing long-range dependencies, processing redundant information efficiently, and extracting high-level semantic concepts. To address these challenges, we propose a novel approach that more accurately reflects human cognition. This paper introduces \textbf{\ModelName}: \ModelNameExtI, featuring two versatile modules that can enhance existing video-language models or operate as a standalone system. Our \SubmoduleIExt\ (\SubmoduleI) efficiently aggregates representations from micro to semi-macro levels, reducing computational overhead while preserving temporal dependencies. Our \SubmoduleIIExt\ (\SubmoduleII) enriches these representations with semantic information by focusing on broader context, dramatically reducing feature dimensionality while preserving relevant macro-level information. We demonstrate that these modules can be seamlessly integrated into existing SOTA models, \textbf{consistently improving their performance while reducing inference latency by up to 43\% and memory usage by 46\%}. As a standalone system, \ModelName\ achieves state-of-the-art performance across multiple long-video understanding benchmarks in both zero-shot and fully-supervised settings. Our project page and code can be found \href{https://joslefaure.github.io/assets/html/hermes.html}{here}.
\end{abstract}    
\section{Introduction}
\label{sec:intro}
Video understanding reflects how humans perceive the world through one of our most essential senses, sight, and drives a wide range of visual and multimodal applications. Whether we want to create better video summarization tools, index and retrieve specifics from the vast and ever-expanding array of video content, or improve content moderation and copyright enforcement, we need models that excel at video understanding. This requirement extends beyond short videos with few frames — a task that image models can already handle adequately — to encompass the analysis of extended video content spanning minutes and comprising thousands of interrelated frames.

Long-form video understanding is challenging for several reasons. First and foremost is the \textit{temporal complexity}, as it requires handling a large number of frames throughout the video. Second, it requires a \textit{semantic understanding of high-level concepts} as well as the narrative structure. The third challenge is the \textit{memory and computational constraints}, making it non-trivial to solve the previous two challenges. Attempts to address these issues have been made by researchers who mainly borrow ideas from short videos~\cite{wu2021towards,miech2020end}, which is a more mature area of research encompassing action recognition and video classification, among others, and for which datasets are more abundant. However, these approaches, which adopt techniques such as pooling \cite{faure2023holistic}, or 3D convolutions \cite{feichtenhofer2019slowfast}, often do not fully account for the unique characteristics of long videos that distinguish them from a simple concatenation of short video segments. Some ideas about short-video modeling, especially for those at the spatial level, may also hold for longer ones, but when it comes to long-term modeling, macro-level representations should be extracted efficiently.

In video understanding, we can distinguish between two types of information: 
\textbf{episodic and semantic}\footnote{We elaborate on this in \Cref{cognition}}. Episodic information refers to specific, sequential events that occur in a video, while semantic information encompasses overarching themes and concepts. To illustrate, consider the scene presented in \Cref{fig:teaser}. Episodic information includes observing adolescents interacting at a baseball game, followed by tense exchanges between a mother and father. These are specific, time-bound events that unfold sequentially. In contrast, semantic information involves recognizing the broader context of youth sports culture and the backdrop of media's influence on domestic life. This high-level understanding concisely overviews the scene and actions, transcending specific moments.

Building on these concepts, we propose \textit{\ModelNameExtI~(\textbf{\ModelName})}, featuring two modular components that can either work together as a complete system or integrate into existing models. The \textbf{E}pisodic \textbf{CO}mpressor (\textbf{\SubmoduleI}) aggregates key contiguous information as we process the video, shaping understanding sequentially while reducing computational overhead and the \textbf{SE}mantic re\textbf{TR}iever (\textbf{\SubmoduleII}) identifies and extracts high-level cues that provide a concise overview of the scene and actions. \ModelName~ achieves state-of-the-art performance on four long-form video understanding benchmarks in both zero-shot and fully-supervised settings, outperforming the state-of-the-art by 7.3\% on LVU\cite{wu2021towards} and 14.9\% on MovieChat-1k~\cite{song2024moviechat}.

Our key contributions are as follows:
\begin{itemize}
    \item We develop a versatile framework for processing and understanding long-form videos that can either operate as a standalone system or enhance existing models through modular integration.
    \item We propose an \SubmoduleIExt\ (\SubmoduleI) that can replace or augment existing memory mechanisms, consistently improving model performance while reducing inference latency and GPU memory usage by up to 43\% and 46\%, respectively.
    \item We develop a \SubmoduleIIExt\ (\SubmoduleII) that enhances video understanding by distilling high-level semantic cues, providing substantial accuracy improvements with minimal computational overhead.
\end{itemize}
Through comprehensive evaluation across multiple benchmarks, detailed ablation studies, and extensive integration experiments with existing SOTA models, we validate the effectiveness of \SubmoduleI\ and \SubmoduleII, demonstrating their complementary roles in enhancing long-form video understanding both as standalone components and as plug-in modules.

\begin{figure*}[ht!]
    \centering
    \includegraphics[width=\textwidth]{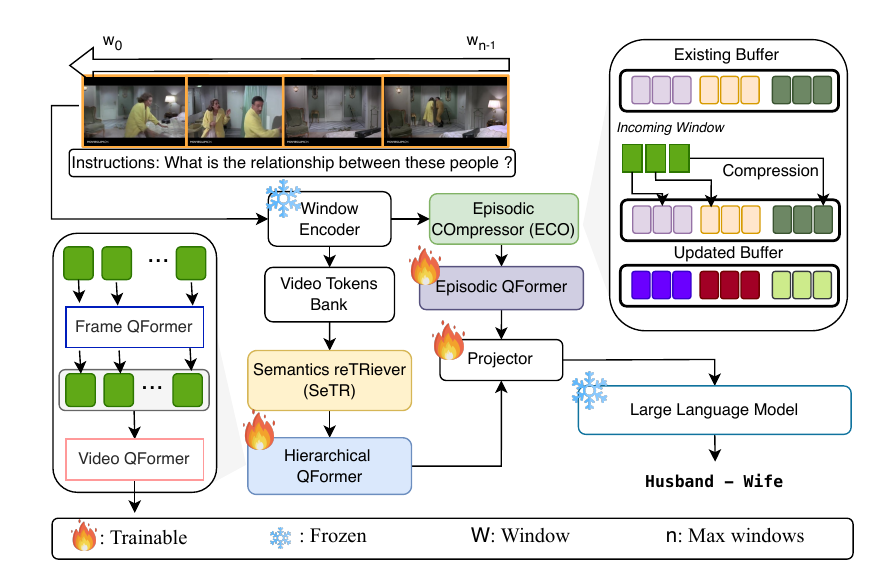}
    \caption{\textbf{\ModelName~framework overview:} We stream through a video window-by-window and extract features using a frozen ViT. Each window feature is processed by the \SubmoduleIExt~(\SubmoduleI) in an online fashion, discarding redundancies along the way and retaining video episodes that are passed to an episodic Q-Former. The video token bank contains the concatenated features of every window, and \SubmoduleII\ selects only the high-level information to pass to a hierarchical frame-to-sequence Q-Former. The episodic and high-level representations are then concatenated before being fed to the frozen LLM, which outputs a text following the instructions.}
    \label{fig:method}
\end{figure*}
\section{Related Work}
\label{sec:relatedwork}

\noindent \textbf{Action recognition} is an essential task in video understanding, primarily focusing on identifying specific actions within short video clips. Various approaches have been developed, with convolutional neural networks forming the core of many of them. Early work by \cite{ji20123d} utilized 3D convolutions, while \cite{varol2017long} employed temporal convolutions. 2D CNNs coupled with temporal modeling have also been explored, with representative works such as Temporal Difference Networks (TDN) \cite{ng2018temporal} and Event Adaptive Networks (EAN) \cite{tian2022ean}. More recently, transformer-based models have gained prominence with works such as \cite{faure2023holistic}, \cite{xu2021long}, and \cite{zhang2022actionformer}.

\noindent \textbf{Video question answering (VideoQA)} aims to answer questions related to video content, requiring a deep understanding of both visual and textual information. Datasets such as ActivityNet-QA \cite{yu2019activitynet} for short videos, and MovieChat-1k for long videos \cite{song2024moviechat} provide benchmarks for evaluating models in this field, allowing for several research endeavors on this subject \cite{zhang2020action,zhuang2020multichannel,pan2023retrieving}. 

\noindent \textbf{Long-form video understanding} presents unique challenges due to the extended duration and complex narrative structures involved. Datasets with these properties include LVU \cite{wu2021towards}, COIN \cite{tang2019coin}, Breakfast \cite{kuehne2014language}, and MovieChat-1k \cite{song2024moviechat}. Traditional approaches to tackling such a task often extend methods designed for short videos to handle longer sequences, such as pooling over the temporal dimension \cite{tang2020asynchronous,faure2023holistic}. Other methods such as \cite{wu2021towards,he2024ma,wu2022memvit} and \cite{song2024moviechat} explore memory techniques via token compression. Additionally, \cite{tian2024coding} introduced a video semantic compression framework using low-level bitrate coding. \cite{wang2023selective} introduced selective structured state-spaces for long-form videos, followed by others \cite{islam2022long,islam2023efficient} exploiting the ability of state-space models to retain long-term context.

\noindent \textbf{Video-Language Models: }Recent advancements in large language models (LLMs) \cite{touvron2023llama,chiang2023vicuna} have piqued researchers' curiosity regarding their use for video understanding \cite{maaz2023video}. It turns out to be a good match, as understanding videos often involves transforming their content into words, whether it's video captioning, video question answering, or even action classification. Frameworks such as \cite{song2024moviechat} and \cite{he2024ma} employ memory techniques to handle extensive video content while \cite{ren2024timechat} presents TimeChat, explicitly conditioning the model to manage time-dependent information.

\section{Problem Statement}
\label{problem}
Given a long video $V = \{f_1, f_2, \ldots, f_N\}$, where $f_i$ represents the $i$-th frame and $N$ is the total number of frames, our objective is to develop a model $\mathbf{M}$ that can efficiently process $V$ and construct an internal understanding $U$ of its content. This understanding should enable the model to answer queries $Q$ or follow instructions $I$ related to the video content.
Formally, we aim to find an optimal function:
\begin{equation}
\mathbf{M} : (V, I) \rightarrow U
\end{equation}
such that:
\begin{itemize}
    \item $U$ captures episodic and semantic information from $V$.
    \item $U$ can be used to maximize the probability $P(A|Q, U)$ of generating correct answers $A$ to queries $Q$ about the video.
\end{itemize}
The key challenges in this formulation are:
\begin{itemize}
    \item[$\bullet$] \textbf{Temporal Complexity:} Efficiently processing $N$ frames, where $N$ can be very large.
    \item[$\bullet$] \textbf{Semantic Understanding:} Extracting high-level concepts and narrative structure from video content.
    \item[$\bullet$] \textbf{Memory Constraints:} Developing a method to maintain relevant information without exhausting computational resources.
\end{itemize}

Addressing these challenges requires an approach that can effectively compress temporal information while preserving both detailed episodic content and high-level semantic understanding. In the following section, we propose a cognitively inspired framework to tackle these challenges.

\section{Proposed Framework: \ModelName}
\label{sec:method}
Our goal is to enhance video understanding by loosely drawing inspiration from human visual processing, rather than developing a new LLM or fine-tuning existing ones. To achieve this, we introduce a method that, given a video and a set of instructions, generates the specified output, such as video question answering (VQA) or video classification. \Cref{fig:method} provides a high-level overview of our framework.

Our approach addresses the challenges identified in Section \ref{problem} through two core principles of human perception:
\begin{enumerate}
\item An \textbf{\SubmoduleIExt\ (\SubmoduleI)}, which structures a video into meaningful segments:
\begin{equation}
\SubmoduleI: \{f_1, f_2, \ldots, f_N\} \rightarrow \{e_1, e_2, \ldots, e_K\}
\end{equation}
where $K \ll N$, and $e_i$ represents compressed episodes.

\item A \textbf{\SubmoduleIIExt\ (\SubmoduleII)}, which extracts high-level semantic context: 
\begin{equation}
\SubmoduleII: \{f_1, f_2, \ldots, f_N\} \rightarrow \{s_1, s_2, \ldots, s_L\}
\end{equation}
where $L \ll N$, and $s_i$ represents extracted semantics.
\end{enumerate}

The final understanding $U$ is generated by combining the outputs of \SubmoduleI\ and \SubmoduleII:
\begin{equation}
U = G(\SubmoduleI(V, I), \SubmoduleII(V))
\end{equation}
where $G$ is a function that integrates episodic and semantic information.

Details on \SubmoduleI\ and \SubmoduleII\ are provided in \Cref{subsec:\SubmoduleI} and \Cref{subsec:\SubmoduleII}, respectively. First, we describe our window encoder approach which serves as the foundation for both components.

\subsection{Window Encoder}
\label{subsec:window}
Our model takes as input a video of arbitrary length. To batch process the video, we first specify a number of frames \( N \) to extract, leading to \( \mathbf{v} = \{\mathbf{f}_1, \mathbf{f}_2, \ldots, \mathbf{f}_N\} \), where \( \mathbf{f}_t \) denotes the \( t \)-th frame. The ViT-G/14 encoder~\cite{fang2023eva} progressively encodes non-overlapping windows of the video data. The window size \( w \) is a divisor of \( N \) and determines how many frames to encode at once. The features of the $k$-th window are denoted as \(\mathcal{W}_{k} \in \mathbb{R}^{B \times w \times T \times C}\), where $B$ is the batch size, \(T\) the number of visual tokens, and \(C\) the number of channels. \(\mathcal{W}_{k}\) are then passed on to the \SubmoduleIExt\ (\SubmoduleI) described in \Cref{subsec:\SubmoduleI}.

\subsection{\SubmoduleI: \SubmoduleIExt}
\label{subsec:\SubmoduleI}
Long videos often contain redundant information, making it crucial to identify and consolidate key episodic elements efficiently. To address this, we propose \SubmoduleI~which maintains a memory buffer with a maximum number of episodes \(E\). Upon receiving a window of frame features, \(\mathcal{W}_{k}\), we first check whether the buffer $\mathcal{M}$ has sufficient bandwidth to support the incoming features. If it does, we simply concatenate them to the buffer; otherwise, we proceed with the compression. At its core, \SubmoduleI\ is a distribution process that determines the episode to which a certain frame belongs. It can be summarized as:
\begin{equation}
\mathcal{M} = 
\begin{cases} 
\mathcal{M} \oplus \mathcal{W}_{k} & \text{if } \|\mathcal{M}\| + \|\mathcal{W}_{k}\| \leq E \\
\text{\SubmoduleI}(\mathcal{M}, \mathcal{W}_{k}) & \text{otherwise}
\end{cases}
\end{equation}
where $\oplus$ is the concatenation operation, $\|\mathcal{M}\|$ and $\|\mathcal{W}_{k}\|$  are the sizes of the buffer and the incoming features, respectively.

\begin{algorithm}[ht!]
\caption{\SubmoduleI: \SubmoduleIExt}\label{alg:submoduleI}
\begin{algorithmic}[1]
\State $\mathcal{A} \gets \mathcal{M} \oplus \mathcal{W}_{k}$
\While{$\|\mathcal{A}\| > E$}
    \State $(i^*, j^*) \gets \arg\max_{i \neq j} \frac{\mathcal{A}_i \cdot \mathcal{A}_j}{\|\mathcal{A}_i\| \|\mathcal{A}_j\|}$
    \State $\mathcal{A}_{i^*} \gets \frac{(\mathcal{A}_{i^*} + \mathcal{A}_{j^*})}{2}$
    \State $\mathcal{A} \gets \mathcal{A} \setminus \mathcal{A}_{j^*}$
\EndWhile
\State $\mathcal{M} \gets \mathcal{A}$
\end{algorithmic}
\end{algorithm}

\SubmoduleI\ works as \Cref{alg:submoduleI} where $\mathcal{M}$ is the existing buffer, \(\mathcal{W}_{k}\) represents the incoming window of frame features, \(\mathcal{A}\) is the concatenated buffer and new window, and $\|\mathcal{A}\|$ is the size of \(\mathcal{A}\). To summarize \SubmoduleI, \(\frac{\mathcal{A}_i \cdot \mathcal{A}_j}{\|\mathcal{A}_i\| \|\mathcal{A}_j\|}\) computes the cosine similarity between frame features \(\mathcal{A}_i\) and \(\mathcal{A}_j\), \(\arg\max_{i \neq j}\) finds the pair of frames with the highest cosine similarity, \(\frac{(\mathcal{A}_{i^*} + \mathcal{A}_{j^*})}{2}\) combines the most similar frames, and \(\mathcal{A} \setminus \mathcal{A}_{j^*}\) removes the frame \(\mathcal{A}_{j^*}\) from \(\mathcal{A}\) after merging. The process repeats until the size of \(\mathcal{A}\) is within the maximally allowed episodes \(E\).

\subsection{Episodic Q-Former}
\label{subsec:eqformer}
To aggregate learned queries into episodes as we did the video features, we integrate \SubmoduleI~as a pruning module within the Q-Former architecture (initialized with weights from \cite{instructblip}). Given initial queries and instructions, we perform self-attention on these queries followed by cross-attention between the queries and visual representations $\mathcal{M}$. The enhanced queries then undergo an \SubmoduleI-like process, where we iteratively merge similar queries across video windows, effectively forming video query episodes of high information density. The following equation summarizes the process,
\begin{equation}
Q = \text{\SubmoduleI}_\mathfrak{q}\left( \text{CA} \left( \text{SA} (Q_0), \mathcal{M} \right) \right)
\end{equation}
where \( Q_0 \) represents the initial queries, \( \mathcal{M} \) denotes the visual representations from the visual \SubmoduleI, \(\text{SA}(Q_0)\) applies self-attention on the initial queries, and \(\text{CA}( \cdot, \mathcal{M})\) performs cross-attention between the self-attended queries and the visual representations. Finally, \(\text{\SubmoduleI}_\mathfrak{q}(\cdot)\) -- note the $\mathfrak{q}$ to differentiate it from the visual \SubmoduleI\ -- applies the iterative merging process similar to the compression detailed in \Cref{subsec:\SubmoduleI} on the queries.
The episodic Q-Former outputs \(Q \in \mathbb{R}^{B \times q \times C'}\) with $B$, $q$ and $C'$ alluding to the batch size, the number of queries and the channel dimension, respectively.

\subsection{\SubmoduleII: \SubmoduleIIExt}
\label{subsec:\SubmoduleII}
To complement \SubmoduleI\ and capture higher-level semantic information from the video, we develop a \SubmoduleIIExt~(\SubmoduleII). \SubmoduleII\ is designed to identify and consolidate important high-level information that may be scattered (contiguously or not) throughout the video. Given a video feature tensor $F \in \mathbb{R}^{B \times N \times T \times C}$, where $B$ is the batch size, $N$ the number of frames, $T$ the number of tokens per frame and $C$ the channel dimension, \SubmoduleII\ operates as follows: we first normalize $F$ to ensure consistent scaling across features. Second, we apply a stride of $k$ to create two groups, group $X$ containing every $k$-th frame, resulting in $\frac{N}{k}$ frames and group $Y$ with the remaining $N - \frac{N}{k}$ frames. Third, we calculate dot product similarity scores between frames in $X$ and $Y$. Finally, for each frame in $Y$, we merge it with its most similar frame in $X$, based on the computed scores by taking their mean.

This process effectively reduces the number of frames from $N$ to $\frac{N}{k}$, consolidating semantic information while maintaining the most relevant features across the video timeline. The resulting semantic representations are denoted as $F' \in \mathbb{R}^{B \times \frac{N}{k} \times T \times C}$. We evaluate the effectiveness of this approach in \Cref{subsec:ablations}. While ToMe~\cite{bolya2022token} have explored token reduction in vision transformers, their approach and objectives differ significantly from ours. Their method focuses on minor token reductions within individual frames, specifically between different layers of a Vision Transformer. In contrast, \SubmoduleII\ retains the most salient frames while significantly reducing redundancies.

\subsection{Hierarchical QFormer}
\label{subsec:hierarchical}
Following our \SubmoduleII, is a hierarchical Q-Former composed of a frame Q-Former ($fQFormer$), a frame-to-sequence adapter and a video Q-Former ($vQFormer$). The frame Q-Former enhances each semantic piece of information, independently of the others, and the video Q-Former consolidates them. The resulting query $Q_{sem} \in \mathbb{R}^{B \times q \times C'}$ contains the semantic representations of the entire video.
\begin{equation}
    Q_{sem} = vQFormer(Linear(fQFormer(F')))
\end{equation}

\subsection{From Representations to Natural Language}
After obtaining the episodic representations $Q$ and the semantic representations $Q_{sem}$, we prepare them for input into a Large Language Model (LLM). Specifically, we concatenate $Q$ and $Q_{sem}$ to form a unified representation vector. This concatenated vector is then projected into the input embedding space of the LLM using a learned linear transformation. In our implementation, we utilize a Vicuna-7B model \cite{chiang2023vicuna} as the LLM. The model, conditioned on this projected representation and guided by task-specific instructions, generates the requested natural language output. This approach allows us to leverage the LLM's pretrained knowledge and language generation capabilities while incorporating our task-specific episodic and semantic information. The process is summarized by the following equation:
\begin{equation}
    \hat{Y} = \text{LLM}(U, I)
\end{equation}
where $U = W[Q; Q_{sem}] + b$ denotes the understanding stemming from the aggregation of semantic and episodic information, $\hat{Y}$ is the generated output, $[Q; Q_{sem}]$ the concatenation of $Q$ and $Q_{sem}$, $W$ and $b$ are the learned projection matrix and bias respectively, and $I$ represents the task-specific instructions.

\begin{table}[b!]
\centering
\begin{tabular}{l|cc|cc}
\toprule
\multirow{2}{*}{\textbf{Model}} & \multicolumn{2}{c|}{\textbf{Global}} & \multicolumn{2}{c}{\textbf{Breakpoint}} \\
\cline{2-5}
 & Acc. & Score & Acc. & Score \\
\midrule
MovieChat~\citep{song2024moviechat} & 63.7 & 3.15 & 48.1 & 2.46 \\
Video-ChatGPT~\citep{maaz2023video} & 58.7 & 2.89 & 47.8 & 2.43 \\
Video-LLaMA~\citep{zhang2023video} & 56.3 & 2.72 & 45.8 & 2.11 \\
VideoChat~\citep{li2023videochat} & 60.2 & 3.08 &  46.3 & 2.32 \\
\midrule
\textbf{\ModelName~(Ours)} & \textbf{78.6} & \textbf{4.23} & \textbf{57.3} & \textbf{3.29} \\
\textcolor{gray}{\textit{\ModelName~(Ours)}\textsuperscript{\textdaggerdbl}} & \textcolor{gray}{\textit{84.9}} & \textcolor{gray}{\textit{4.40}} & \textcolor{gray}{\textit{65.8}} & \textcolor{gray}{\textit{3.65}} \\
\bottomrule
\end{tabular}
\caption{\textbf{Zero-shot performance on MovieChat-1k}. Our model significantly outperforms existing methods. The model marked with \textsuperscript{\textdaggerdbl} is fully supervised.}
\label{tab:moviechat}
\end{table}

\begin{table*}[!t]
\centering
\centering
\begin{tabular}{l|ccc|cccc|c|c|c}
\toprule
& \multicolumn{8}{c|}{\textbf{LVU}} & & \\ 
\cline{2-9}
\multirow{1}{*}{\textbf{Model}} & \multicolumn{3}{c|}{\textbf{Content}} & \multicolumn{4}{c|}{\textbf{Metadata}} & \multirow{2}{*}{\textbf{Avg}} & \multirow{1}{*}{\textbf{Breakfast}} & \multirow{1}{*}{\textbf{COIN}} \\
\cline{2-8}
 & Relation & Speak & Scene & Director & Genre & Writer & Year & & & \\
\midrule
FACT \citep{Lu_2024_CVPR} & - & - & - & - & - & - & - & - & 86.1 & - \\
Obj. Transformer~\citep{wu2021towards} & 53.1 & 39.4 & 56.9 & 52.1 & 54.6 & 34.5 & 39.1 & 47.1 & - & - \\
VIS4mer~\citep{islam2022long}  & 57.1 & 40.8 & 67.4 & 62.6 & 54.7 & 48.8 & 44.8 & 53.7 & 88.2 & 88.4 \\
TranS4mer~\citep{islam2023efficient} & 59.5 & 39.2 & 70.9 & 63.9 & 55.9 & 46.9 & 45.5 & 54.5 & 90.3 & 89.2 \\
S5~\citep{wang2023selective} & 67.1 & 42.1 & 73.5 & 67.3 & \underline{65.4} & 51.3 & 48.0 & 59.2 & 90.7 & 90.8 \\
Movies2Scenes~\citep{Chen_2023_CVPR} & \textbf{71.2} &  42.2 & 68.2 & 70.9 & 57.8 & 55.9 & \underline{53.7} & 60.0 & - & - \\
MA-LMM~\citep{he2024ma} & 58.2 & \underline{44.8} & \underline{80.3} & \underline{74.6} & 61.0 & \underline{70.4} & 51.9 & \underline{63.0} & \underline{93.0} & \underline{93.2} \\
\midrule
\textbf{\ModelName~(Ours)} & \underline{67.6} & \textbf{47.5} & \textbf{90.0} & \textbf{82.6} & \textbf{69.5} & \textbf{77.2} & \textbf{57.7} & \textbf{70.3} & \textbf{95.2} & \textbf{93.5} \\
\bottomrule
\end{tabular}
\caption{\textbf{SOTA Comparison on the LVU, Breakfast and COIN datasets:} The table presents Top-1 accuracy for various models. Unlike the minor incremental improvements observed among other methods, our model demonstrates a significant performance leap, outperforming its nearest competitor by 7.3\% on LVU and 2.2\% on Breakfast. The highest score is highlighted in \textbf{bold}, and the second highest is \underline{underlined}.}
\label{tab:lvu}
\end{table*}

\begin{figure*}[t!]
    \begin{minipage}[b!]{0.45\textwidth}
        \centering
        \begin{tabular}{lccc}
        \toprule
        Model & Acc. & Time & Mem. (GB) \\
        \midrule
        LongVA (7B) & 54.11 & 1 & 42.5 \\
        \midrule
        + \SubmoduleI & 54.19 & 0.700 \textcolor{ForestGreen}{\scriptsize (-30\%)} & \textbf{22.9} \\
        + \SubmoduleII & \textbf{54.56}  & 0.726 \textcolor{ForestGreen}{\scriptsize (-27\%)} & 32.7 \\
        \bottomrule
        \end{tabular}
        \captionof{table}{Zero-shot performance comparison of LongVA with and without \SubmoduleI\ and \SubmoduleII\ integration on VideoMME.}
        \label{tab:submodules-longva}
    \end{minipage}%
    \hfill
    \begin{minipage}[b!]{0.48\textwidth}
        \centering
        \begin{tabular}{lccc}
        \toprule
        Model & Acc. & Time & Mem. (GB) \\
        \midrule
        LLaVA-OV (7B) & 58.26 & 1 & 40.6 \\
        \midrule
        + \SubmoduleI & 58.93 & 0.650 \textcolor{ForestGreen}{\scriptsize (-35\%)} & \textbf{33.4} \\
        + \SubmoduleII & \textbf{59.30}  & 0.673 \textcolor{ForestGreen}{\scriptsize (-33\%)} & \textbf{33.4} \\
        \bottomrule
        \end{tabular}
        \captionof{table}{Zero-shot performance comparison of LLaVA-OneVision with and without \SubmoduleI\ and \SubmoduleII\ integration on VideoMME.}
        \label{tab:submodules-llavaov}
    \end{minipage}
\end{figure*}

\begin{table}[ht]
    \centering
    \begin{tabular}{lcccc}
        \toprule
        Model & Acc. & Score & Time & Mem. (GB) \\
        \midrule
        MA-LMM & 73.3 & 4.05 & 1 & \textbf{30.2} \\
        \midrule
        + \SubmoduleI & 76.7 & 4.14 & 0.569 \textcolor{ForestGreen}{\scriptsize (-43\%)} & \textbf{30.2} \\
        + \SubmoduleII & \textbf{77.1}  & \textbf{4.16} & 1.015 \textcolor{magenta}{\scriptsize (+1.5\%)} & 32.5 \\
        \bottomrule
        \end{tabular}
        \captionof{table}{Zero-shot comparison of MA-LMM with and without \SubmoduleI\ as memory manager and integrating \SubmoduleII\ on MovieChat-1k.}
        \label{tab:submodules-malmm}
\end{table}

\section{Experiments}
\label{sec:experiments}

\subsection{Datasets and Evaluation Metrics}
\label{subsec:dataset}

We evaluate our approach on two primary tasks: long-form video classification and long-form video question answering.

For long-form video classification, we utilize three datasets. The first, LVU \cite{wu2021towards}, focuses on movie content, offering a rich source of narrative and thematic video data. The second, Breakfast \cite{tang2019coin}, consists of instructional videos that emphasize procedural understanding. Lastly, COIN \cite{kuehne2014language} is another instructional video dataset that covers a wider range of procedural activities compared to Breakfast. We report top-1 classification accuracy on these datasets.

For long-form video question answering, we employ the MovieChat-1k dataset \cite{song2024moviechat} and report both zero-shot and fully-supervised results. As evaluation metrics, we follow the evaluation protocol developed by \cite{maaz2023video}, employing GPT-3.5-turbo \cite{brown2020language} to assess both accuracy and answer quality score. We also perform plug-and-play analysis of \SubmoduleI~and \SubmoduleII~on three SOTA methods including MA-LMM \cite{he2024ma}, LongVA \cite{zhang2024long} and LLaVA-OneVision \cite{li2024llava} and show enhanced performance on VideoMME \cite{fu2024video} and MovieChat-1k.

\subsection{Quantitative Results}
For VQA, we evaluate on the MovieChat-1k dataset \cite{song2024moviechat}. As shown in \Cref{tab:moviechat}, \ModelName\ surpasses recent LLM-based models including MovieChat \cite{song2024moviechat}, Video-ChatGPT \cite{maaz2023video}, Video-LLaMA \cite{zhang2023video}, and VideoChat \cite{li2023videochat}, achieving a substantial 14.9\% improvement over previous best results.
On standard long-form video classification benchmarks including LVU \cite{wu2021towards}, Breakfast \cite{kuehne2014language}, and COIN \cite{tang2019coin}, \ModelName\ consistently outperforms existing approaches (\Cref{tab:lvu}). We compare our model against three categories of methods: transformer-based models \cite{wu2021towards, Chen_2023_CVPR, Lu_2024_CVPR}, hybrid architectures combining state-space and transformer approaches \cite{islam2022long, islam2023efficient, wang2023selective}, and the LLM-based model MA-LMM \cite{he2024ma}. Notably, \ModelName\ achieves a 7.3\% improvement over the previous state-of-the-art on LVU.

\begin{figure*}[t!]
    \centering
    \begin{minipage}[ht!]{0.3\textwidth}
        \centering
        \begin{tabular}{lcc}
        \toprule
         & Acc. & Score \\
        \midrule
        w/o &  55.1 & 3.55 \\
        Rand. & 76.9 & 4.13 \\
        FIFO &  77.1 & 4.15 \\
        \textbf{\SubmoduleI} &  \textbf{78.6} & \textbf{4.23} \\
        \bottomrule
        \end{tabular}
        \captionof{table}{Ablations on the memory update design of our \SubmoduleIExt.}
        \label{tab:\SubmoduleI}
    \end{minipage}
    \hfill
    \begin{minipage}[ht!]{0.3\textwidth}
        \centering
        \begin{tabular}{lcc}
        \toprule
         & Acc. & Score \\
        \midrule
        w/o & 73.3 & 4.09 \\
        MaxPool & 70.4 & 3.99 \\
        AvgPool & 73.3 & 4.04 \\
        K-Means & 75.7 & 4.11 \\
        \textbf{\SubmoduleII} &  \textbf{78.6} & \textbf{4.23} \\
        \bottomrule
        \end{tabular}
        \captionof{table}{Ablations on different semantic compression methods.}
        \label{tab:\SubmoduleII}
    \end{minipage}
    \hfill
    \begin{minipage}[ht!]{0.3\textwidth}
        \centering
        \begin{tabular}{lcc}
        \toprule
         & Acc. \\
        \midrule
        $fQFormer$ & 93.2 \\
        $vQFormer$ & 94.1 \\
        $HQFormer$ & \textbf{95.2} \\
        \bottomrule
        \end{tabular}
        \captionof{table}{Performance comparison between frame Q-Former, video Q-Former and our hierarchical Q-Former.}
        \label{tab:hierarchical}
    \end{minipage}
\end{figure*}

\begin{figure*}[t!]
    \begin{minipage}[ht!]{0.3\textwidth}
        \centering
        \includegraphics[width=\textwidth]{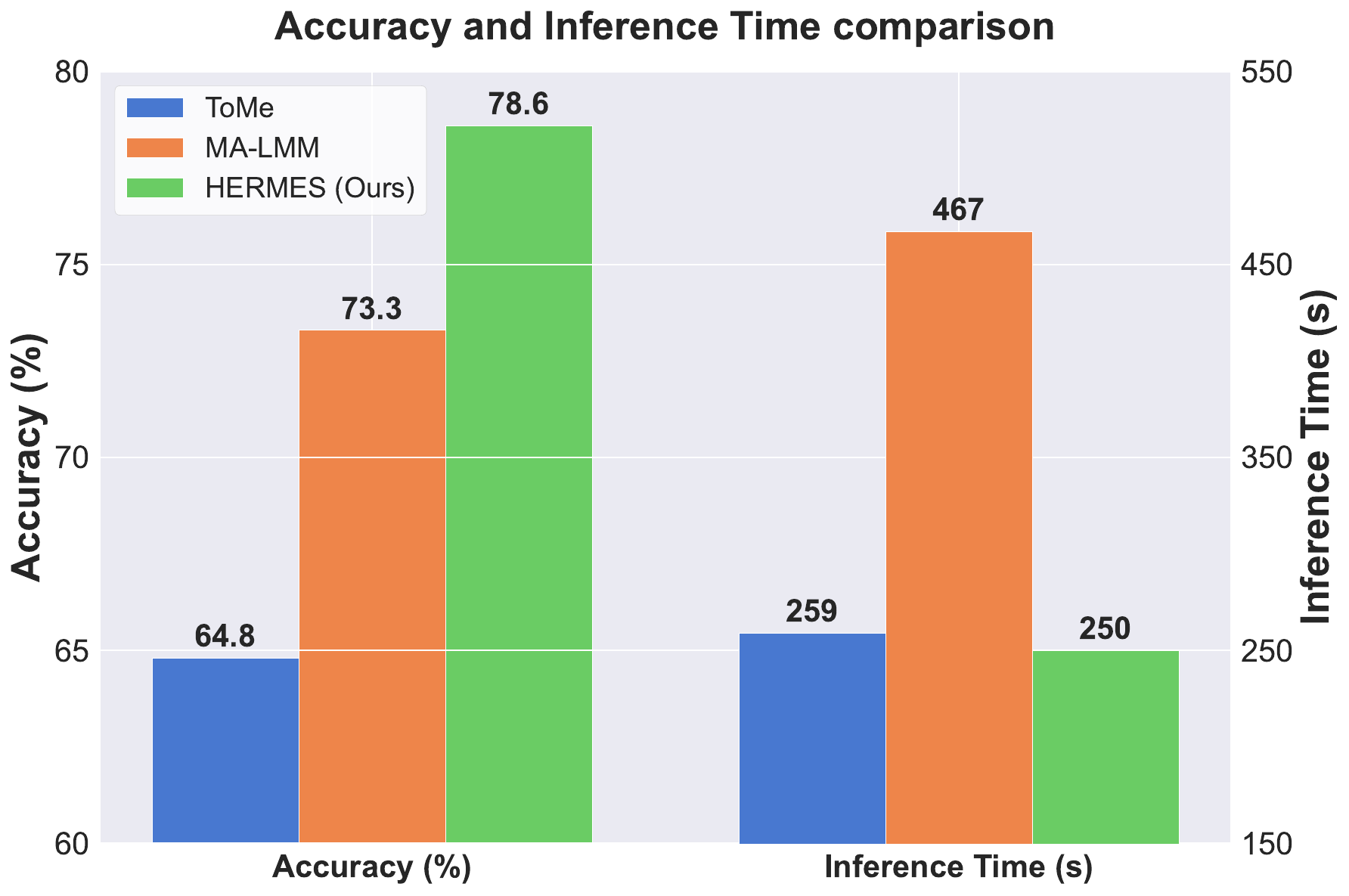}
        \captionsetup{width=0.95\textwidth}
        \caption{Our method is 46\% faster than MA-LMM while being 5.3\% more accurate, and registers an absolute gain of 13.8\% accuracy compared to ToMe.}
        \label{fig:speed}
    \end{minipage}
    \hfill
    \begin{minipage}[ht!]{0.3\textwidth}
        \centering
        \includegraphics[width=\textwidth]{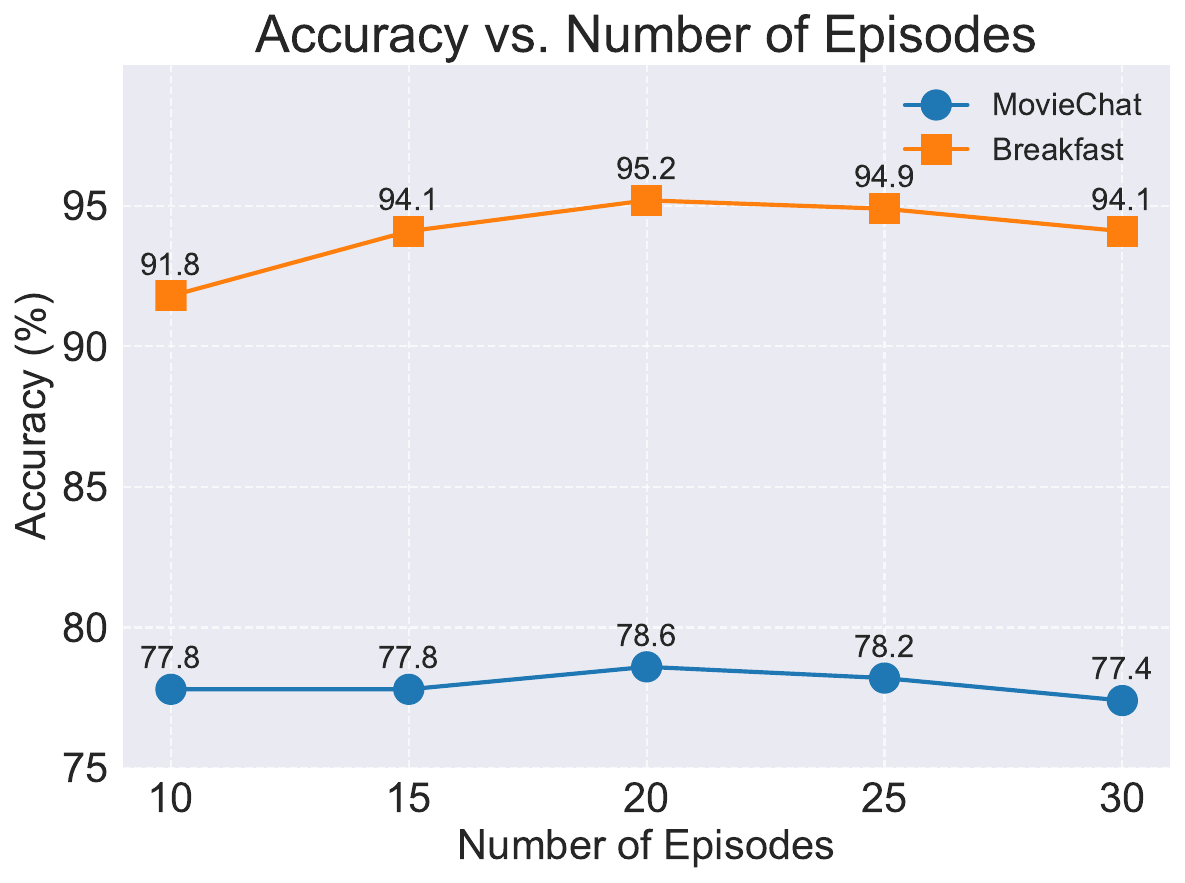}
        \captionsetup{width=0.95\textwidth}
        \caption{Effect of the number of \SubmoduleI\ episodes on the model's accuracy on the MovieChat-1k and Breakfast datasets.}
        \label{fig:acc\SubmoduleI}
    \end{minipage}
    \hfill
    \begin{minipage}[ht!]{0.3\textwidth}
        \centering
        \includegraphics[width=\textwidth]{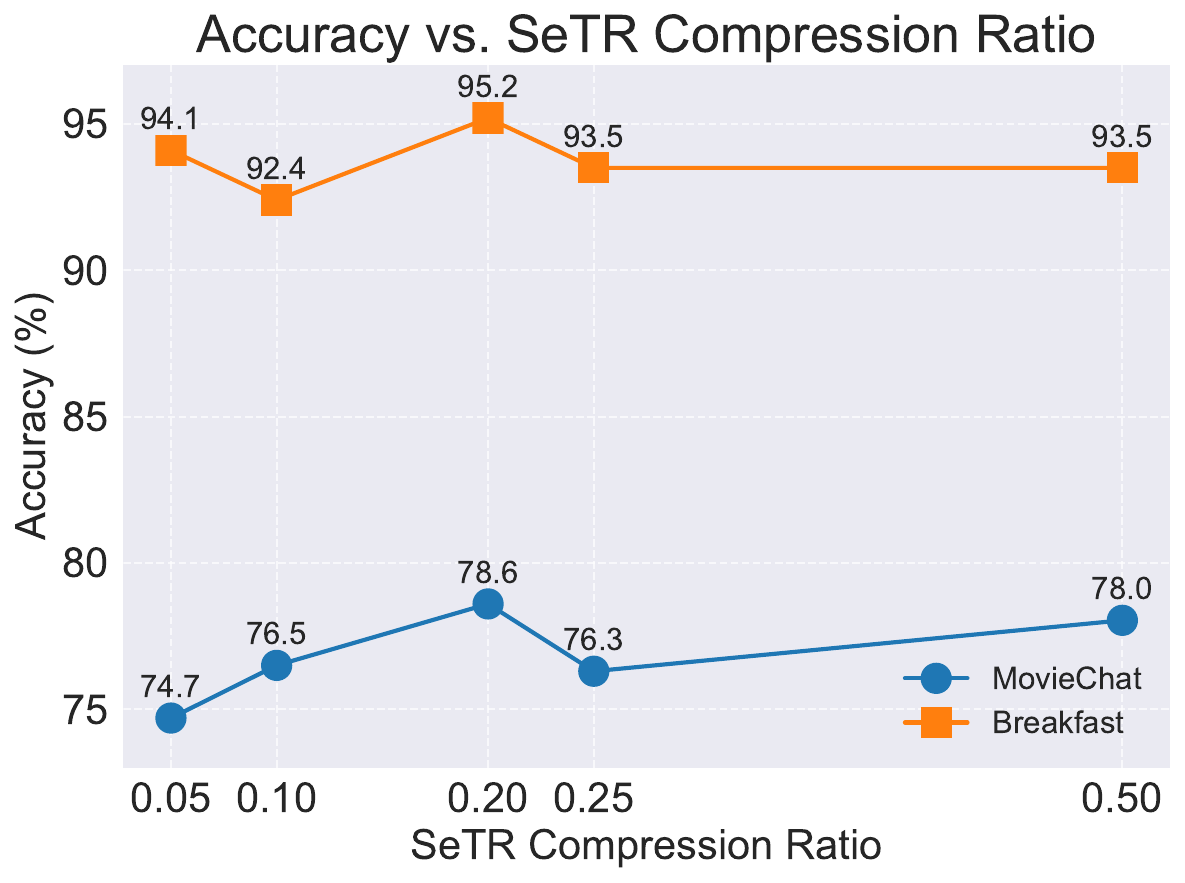}
        \captionsetup{width=0.95\textwidth}
        \caption{Effect of the \SubmoduleII~'s keep ratio on the model's accuracy on the MovieChat-1k and Breakfast datasets.}
        \label{fig:acc\SubmoduleII}
    \end{minipage}
\end{figure*}

\subsection{Pilot Study: \SubmoduleI~and \SubmoduleII~as plug-in modules}
In this pilot study, we demonstrate the versatility and effectiveness of our \SubmoduleIExt\ (\SubmoduleI) and \SubmoduleIIExt~(\SubmoduleII) by integrating them into three state-of-the-art video-language models: MA-LMM \cite{he2024ma}, LongVA \cite{zhang2024long}, and LLaVA-OneVision \cite{li2024llava} and evaluate them on two challenging benchmarks: MovieChat-1k \cite{song2024moviechat} and VideoMME (w/o sub.) \cite{fu2024video}. The results are reported in Tables \ref{tab:submodules-malmm}, \ref{tab:submodules-longva} and \ref{tab:submodules-llavaov}.

\noindent\textbf{\SubmoduleI: A Lightweight Episode Compressor.}
Integrating \SubmoduleI, we notice consistent and substantial improvements across all models. Most notably, replacing MA-LMM's memory bank with \SubmoduleI~yields a significant 3.4\% increase in accuracy while simultaneously reducing inference time\footnote{Inference time relative to the baseline} by 43\% and keeping memory usage constant. Similar efficiency gains are observed when integrated with LongVA \cite{zhang2024long} and LLaVA-OneVision \cite{li2024llava}, where \SubmoduleI~maintains or improves accuracy while reducing latency by 30\% and 35\%, respectively, and almost halving the GPU memory usage in the case of LongVA.

\noindent\textbf{\SubmoduleII: An Efficient Semantics Retriever.}
To further validate the complementary nature of our modules, we integrate \SubmoduleII into the same three models. As evidenced in Tables \ref{tab:submodules-malmm}, \ref{tab:submodules-longva}, and \ref{tab:submodules-llavaov}, \SubmoduleII~consistently enhances the models' performance. For MA-LMM, we observe a substantial 3.8\% increase in accuracy and a 0.11 improvement in score, achieved with only a minimal 1.5\% increase in inference time. This demonstrates \SubmoduleII's ability to extract rich semantic information while maintaining computational efficiency. When combined with \SubmoduleI, the integration of \SubmoduleII~into LongVA and LLaVA-OneVision yields further accuracy improvements of 0.37\% for both models, while preserving the significant latency and memory usage reductions.

The consistent performance improvements achieved by both modules across different architectures and datasets underscore their effectiveness as plug-and-play solutions for enhancing video-language models. For MA-LMM, we evaluate \SubmoduleII~in conjunction with their existing memory bank, demonstrating its ability to extract complementary semantic information. For LongVA and LLaVA-OneVision, we showcase the additive benefits of incorporating both modules sequentially, highlighting their synergistic relationship in improving model capabilities while maintaining efficiency.

\subsection{Ablation Studies}
\label{subsec:ablations}
Ablations are conducted on the MovieChat-1k test set (global mode) using the zero-shot setting with additional ablations on the Breakfast dataset using the fully-supervised setting. These experiments focus on our two primary contributions, \SubmoduleI\ and \SubmoduleII. For extended and more comprehensive ablations, please refer to \Cref{supp:extabl} (in the Supp.). We also visualize the features extracted by each module in \Cref{supp:ecoviz}

\noindent\textbf{How important is \SubmoduleI?} In \Cref{tab:\SubmoduleI}, we demonstrate the critical role of \SubmoduleI\ through several experiments. The results indicate that the absence of our \SubmoduleI\ and the Episodic Q-Former leads to a significant degradation in model performance due to the model lacking micro-level continuous representations. We further explore alternative update strategies, including randomly selecting features to retain (Rand.) and employing a first-in-first-out (FIFO) streaming approach. Our proposed update strategy outperforms both the Rand. and FIFO methods, highlighting its efficacy in retaining more relevant episodes.


\noindent\textbf{How important is \SubmoduleII?} \SubmoduleII\ is designed to complement the episodic knowledge of our model with semantic insights. In \Cref{tab:\SubmoduleII}, we observe that removing \SubmoduleII\ results in a 5\% drop in accuracy. Additionally, we show that naive methods such as max and average pooling are not as effective.

\noindent\textbf{Do we need a \textit{hierachical} Q-Former?} Yes. We conducted an ablation study on the Breakfast dataset \cite{kuehne2014language}, to evaluate the efficacy of our proposed hierarchical Q-Former architecture. As shown in Table \ref{tab:hierarchical}, our hierarchical Q-Former achieves superior performance with an accuracy of 95.2\%, outperforming both flat frame-level ($fQFormer$) and video-level ($vQFormer$). This improvement can be attributed to the hierarchical structure's ability to capture multi-scale features, effectively aggregating information from frame to video level. By first processing frame-level details and then aggregating them at the video level, our approach mitigates information loss that may occur in direct video-level processing while avoiding the computational intensity of processing every frame individually.

\noindent\textbf{How effective and efficient is \SubmoduleI\ compared to other memory compressors?}
To demonstrate the effectiveness and efficiency of our proposed \SubmoduleI, we conduct a comparative analysis against two strong existing compression techniques: ToMe \cite{bolya2022token} and MA-LMM \cite{he2024ma} in \Cref{fig:speed}. We calculate the inference time for each model on the MovieChat-1k dataset. Powered by \SubmoduleI, \ModelName~achieves the highest accuracy (78.6\%) among all models, outperforming MA-LMM by 5.3\% and ToMe by a substantial 13.8\%. \ModelName\ also achieves the highest inference speed among the compared models, while maintaining superior accuracy. It is slightly faster than ToMe and significantly outperforms MA-LMM, reducing inference time by 46\% relative to the latter. These results demonstrate our model's ability to deliver state-of-the-art accuracy without compromising on efficiency.

\noindent\textbf{Hyperparameters for \SubmoduleI\ and \SubmoduleII.} Our experiments on the MovieChat-1k (zero-shot) and Breakfast (fully-supervised) datasets reveal compelling insights into the optimal configuration of \SubmoduleI\ (Figure \ref{fig:acc\SubmoduleI}) and \SubmoduleII\ (Figure \ref{fig:acc\SubmoduleII}). For \SubmoduleI, we discover that an episodic memory size of 20 consistently yields peak performance across both datasets, achieving a 78.6\% accuracy on MovieChat-1k and a 95.2\% on Breakfast. This sweet spot balances comprehensive video representation with computational efficiency, as larger memory sizes show diminishing returns. \SubmoduleII's performance proved equally intriguing, with a keep ratio of 20\% (reducing representations by 80\%) emerging as the optimal choice for both datasets. Such results demonstrate the resilience of \textit{\ModelName} to hyperparameter variations suggesting that it is suitable for deployment across diverse video understanding datasets with minimal hyperparameter tuning.

\subsection{Qualitative Results}
\label{qual}
\begin{figure}[ht!]
    \centering
    \includegraphics[width=0.5\textwidth]{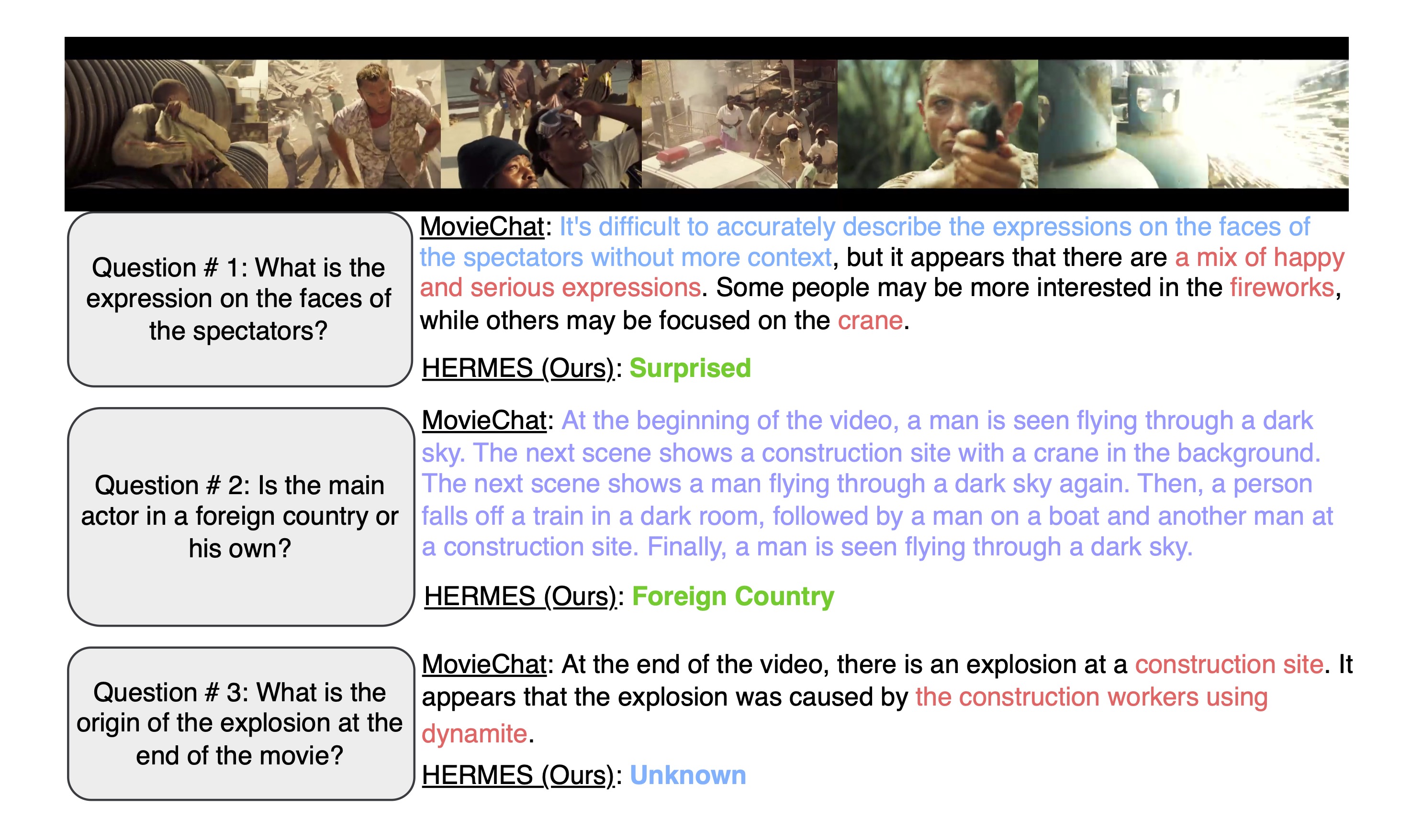}
    \caption{\textbf{Qualitative Results:} We select a challenging video from the MovieChat-1k dataset and pose various difficult questions to both MovieChat \cite{song2024moviechat} and \ModelName. The results demonstrate our model's superior ability to answer both fine-grained questions (Q1 and Q3) and general questions (Q2). Answers highlighted in blue denote tentative answers, red denote wrong answers, purple denote hallucinations, and green denote correct answers.}
    \label{fig:qualitative}
\end{figure}

We present qualitative results on a challenging movie scene from the MovieChat-1k dataset (\Cref{fig:qualitative}) to evaluate our model's capability in answering both fine-grained and general questions about an extended video (14k frames). To rigorously assess the models, we bypass the original Q\&As from the dataset (e.g., Q: What's the time in the video? A: Day, ...) and ask questions that require a deeper understanding of the scene. Our model accurately responds to these questions while exhibiting a candid acknowledgment of its limitations (e.g., Q3). In contrast, MovieChat \cite{song2024moviechat} frequently generates hallucinated and incorrect answers. \ModelName~achieves this performance by processing only 100 out of the 14k frames (approximately 0.7\%), whereas MovieChat processes 2,048 frames, more than 20 times the data utilized by \ModelName. We provide additional qualitative results and failure cases in the supplemetary material, \Cref{appendix:qual} and \Cref{sub:error}.

\section{Limitations}
\label{sec:limitations}
While \ModelName~ demonstrates significant efficiency and performance gains, it relies on heuristics for both episodic compression and semantic retrieval, which may occasionally fail to capture subtle but important temporal details or contextual nuances. Furthermore, the episodic compressor and semantic retriever operate independently, potentially allowing redundancy. Due to computational constraints, we were unable to pretrain \ModelName~ on large-scale video datasets, limiting direct comparisons with extensively pretrained models like LLaVA-OneVision on benchmarks such as VideoMME. Nevertheless, the substantial improvements achieved through our lightweight integration approach suggest promising directions when combined with more computational resources.
\section{Conclusion}
\label{sec:conclusion}
We present \textbf{\ModelName}, a method for enhancing long-form video understanding through two powerful, modular components inspired by cognitive processes. The \SubmoduleIExt\ (\SubmoduleI) captures representations as sequences of continuous actions while significantly reducing computational overhead, and the \SubmoduleIIExt\ (\SubmoduleII) serves as an efficient semantic enrichment mechanism. As standalone components, these modules can be seamlessly integrated into existing video-language models, consistently improving their performance while reducing inference latency. As a complete system, \ModelName\ achieves state-of-the-art results across several long-video datasets, significantly outperforming existing methods. Through extensive experiments on five datasets, and integration studies with three SOTA models, we have demonstrated the effectiveness, efficiency, and versatility of our approach. As model sizes continue to increase and inference efficiency becomes a critical bottleneck in video understanding, our work provides a timely and foundational approach for both enhancing existing LLM-based systems and developing more scalable standalone solutions.
{
    \small
    \bibliographystyle{ieeenat_fullname}
    \bibliography{main}
}

\clearpage
\renewcommand{\thesection}{\Alph{section}}

\maketitlesupplementary

\section{Supplementary Material}
This Supplementary document is organized as follows:
\begin{itemize}
    \item \hyperref[reproducible]{\textbf{A.1} Reproducibility Statement}
    \item \hyperref[impl]{\textbf{A.2} Implementation Details}
    \item \hyperref[model_details]{\textbf{A.3} Model Details}
    \item \hyperref[supp:extabl]{\textbf{A.4} Extended Ablations}
    \item \hyperref[supp:vs]{\textbf{A.5} HERMES vs. MA-LMM vs. MovieChat}
    \item \hyperref[appendix:latency]{\textbf{A.6} A Note on Latency}
    \item \hyperref[appendix:qual]{\textbf{A.7} More Qualitative Results}
    \item \hyperref[sub:error]{\textbf{A.8} Error Analysis: When does \ModelName\ fail and why?}
    \item \hyperref[cognition]{\textbf{A.9} How is our approach related to cognitive processes?}
\end{itemize}

\subsection{Reproducibility Statement}
\label{reproducible}
To facilitate the reproducibility of our work, we will make our code, pretrained models, default hyperparameters, and preprocessed annotations publicly available. Detailed hyperparameters for each dataset are also provided in Table \ref{tab:hyperparameters}. Our model demonstrates efficient performance, completing inference on the MovieChat-1k test set in 13 minutes (22 FPS) using a single V100 GPU (32 GB), and training on the MovieChat-1k dataset in less than 12 minutes with 8x 32 GB GPUs. In contrast to recent LLM-based approaches that necessitate extensive and costly multi-stage pretraining on increasingly large datasets, our model is designed for accessibility, thereby lowering the barrier for researchers without access to high-end computing resources. We achieve high performance while maintaining accessibility by leveraging existing pretrained weights and implementing our training-free \SubmoduleI\ and \SubmoduleII, resulting in a model where finetuning is optional. We also demonstrate the applicability of our modules to existing video models, and are planning to submit pull requests to integrate our modules into these models.

For fully-supervised results, QFormers and adapter are fine-tuned on the respective dataset's training split. For plug-in experiments, ECO and SeTR are inserted into target architectures at inference time, with \textbf{zero additional training}, demonstrating true plug-and-play capability.

\begin{table*}[ht]
    \centering
    \begin{tabular}{lcccccc}
        \toprule
        \textbf{Dataset} & \textbf{Max Epochs} & \textbf{LR} & \textbf{Batch} & \textbf{Frames (N)} & \textbf{Episodes} & \textbf{Keep Ratio} \\
        \midrule
        MovieChat-1k (G) & 1 & 1e-4 & 32 & 100 & 20 & 0.2 \\
        MovieChat-1k (B) & 1 & 1e-4 & 32 & 40 & 10 & 0.5 \\
        LVU & 20 & 1e-4 & 32 & 100 & 20 & 0.2 \\
        COIN & 20 & 1e-4 & 32 & 100 & 20 & 0.2 \\
        Breakfast & 20 & 1e-4 & 32 & 100 & 20 & 0.2 \\
        VideoMME (LongVA) & - & - & 1 & 128 & 32 & 0.125 \\
        VideoMME (Llava-OV) & - & - & 1 & 128 & 32 & 0.125 \\
        \bottomrule
    \end{tabular}
    \caption{Hyperparameters used for different datasets.}
    \label{tab:hyperparameters}
\end{table*}

\subsection{Implementation Details}
\label{impl}
To ensure the reproducibility of our results, we provide training and inference details in \Cref{tab:hyperparameters}. These settings are mostly consistent across different datasets. In the table, LR is the learning rate, and Keep Ratio is the \SubmoduleII\ keep ratio. Episodes refer to the number of episodes to which we compress the input frames (i.e., the capacity of \SubmoduleI). The number of frames (N) represents the quantity of frames retained from the original video to serve as input to the model. These frames are selected by applying a regular stride over the original video's frame sequence, where the stride length is determined by the ratio of original frame count to N. \textit{Max Epoch = 20} means we run the program for 20 epochs, performing evaluation after each epoch, and then pick the model with the highest validation accuracy. MovieChat-1k (G) and MovieChat-1k (B) denote global and breakpoint modes, respectively. All models were trained on 8 V100 GPUs (32GB VRAM each). We test on VideoMME using the zero-shot setting by applying our modules to two different models, the same parameters were used across models for consistency.

\begin{table*}[ht]
    \centering
    \begin{minipage}[ht!]{0.45\textwidth}
        \centering
        \includegraphics[width=\textwidth]{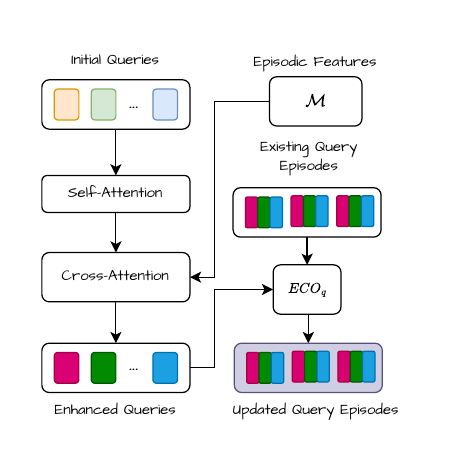}
        \captionof{figure}{\textbf{Illustration of our Episodic QFormer: } We insert our \SubmoduleI~ in the original QFormer to effectively and efficiently compute and aggregate queries across long video sequences. It returns query episodes representing the whole video.}
        \label{fig:eco_q}
    \end{minipage}  
    \hfill
    \begin{minipage}[ht!]{0.45\textwidth}
        \centering
        \includegraphics[width=\textwidth]{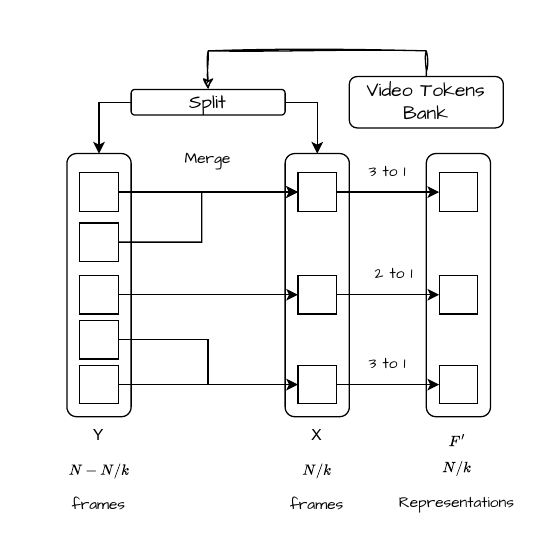}
        \captionof{figure}{\textbf{Illustration of \SubmoduleII:} Our \SubmoduleIIExt~ uses a stride of $k$ split the videos into groups $X$ of $N/k$ frames and $Y$ of $N-\frac{N}{k}$ frames, then merge each frame from $Y$ to its most semantically similar in $X$.}
        \label{fig:setr}
    \end{minipage}  
\end{table*}

\subsection{Model Details}
\label{model_details}
\subsubsection{Details of our Episodic QFormer}
The Episodic Q-Former, as visualized in \Cref{fig:eco_q}, extends the original QFormer architecture by inserting the \SubmoduleIExt~(\SubmoduleI) described in \Cref{subsec:\SubmoduleI}. It begins with a set of initial queries that undergo a self-attention process, enhancing internal query representations. These queries then interact with episodic visual features through cross-attention, allowing the incorporation of contextual visual information. The resulting enhanced queries are fed into our \SubmoduleI~ module alongside existing query episodes, which represent previously processed queries grouped into episodes. \SubmoduleI~ iteratively updates the query episodes, adding the new queries to the existing episodes. This Episodic QFormer allows the model to better handle long sequences or repeated queries by maintaining richer contextual knowledge across iterations.

To mitigate \textit{temporal confusion} during merging, we apply positional encoding (PE) to frame features before ECO. This effectively discourages out-of-order merges by embedding temporal locality directly into similarity calculations. As an ablation, \textbf{removing PE reduces MovieChat-1k accuracy from 78.6 to 77.3} on MovieChat-1k, indicating its effectiveness in preserving temporal coherence despite compression. Other studies such as Transformer-XL \cite{dai2019transformerxlattentivelanguagemodels} and Compressive Transformer \cite{rae2019compressivetransformerslongrangesequence}, also report performance drops when positional biases are removed from their compression modules.

\SubmoduleI~\textbf{implicitly captures event frequency}: frequent events naturally occur across multiple frames and thus have higher likelihoods of being retained or merged into reinforced prototypes within the memory bank. This self-reinforcing mechanism ensures high-importance (and often high-frequency) events remain well-represented. Explicit event frequency tracking is an idea worth exploring, however, we believe it would be more computationally intensive and may force important but infrequent representations out of memory.

\subsubsection{Details of \SubmoduleII}
We design \SubmoduleII~ as an efficient tool to retrieve semantic information from a long video. Given tokens extracted from a long video sequence, we use a stride of size $k$, to form a group of $\frac{N}{k}$ frames representing the number of semantics we want to extract. We then compress the remaining $N-\frac{N}{k}$ frames into extracted $\frac{N}{k}$ frames to obtain the semantic representations. \SubmoduleII~ is illustrated in \Cref{fig:setr}.

\subsection{Extended Ablations}
\label{supp:extabl}
\begin{table*}[ht]
    \centering
    \begin{minipage}[ht!]{0.47\textwidth}
    \captionsetup{width=.9\linewidth}
        \centering
        \includegraphics[width=\textwidth]{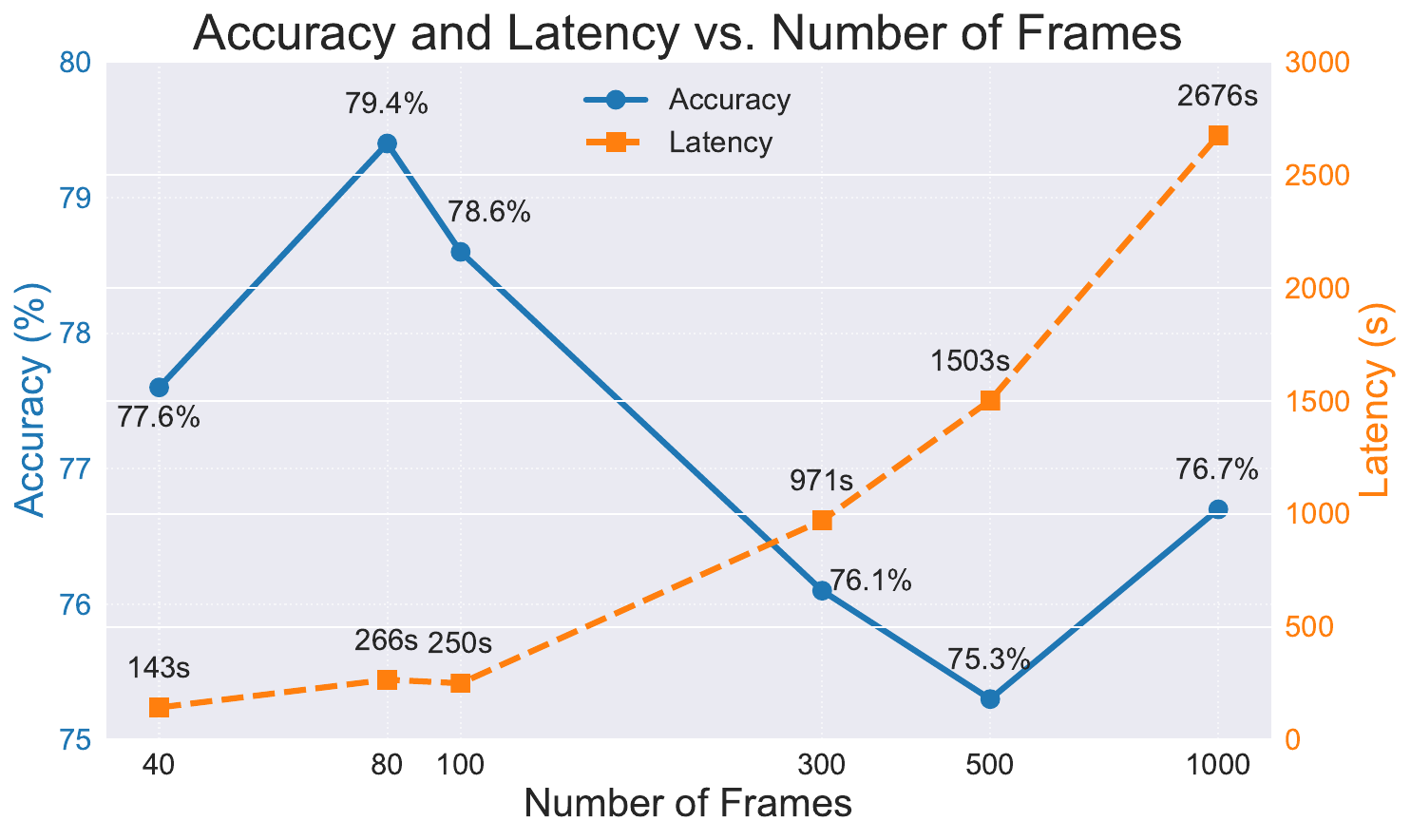}
        \captionof{figure}{\textbf{Accuracy and latency as functions of the number of frames processed: } This figure demonstrates the non-monotonic relationship between accuracy and frame count, with peak performance at 80 frames. Latency increases super-linearly with frame count while accuracy stalls, highlighting the redundancy of video data.}
        \label{fig:acc_latency_frames}
    \end{minipage}  
    \begin{minipage}[ht!]{0.47\textwidth}
    \captionsetup{width=.9\linewidth}
        \centering
        \includegraphics[width=\textwidth]{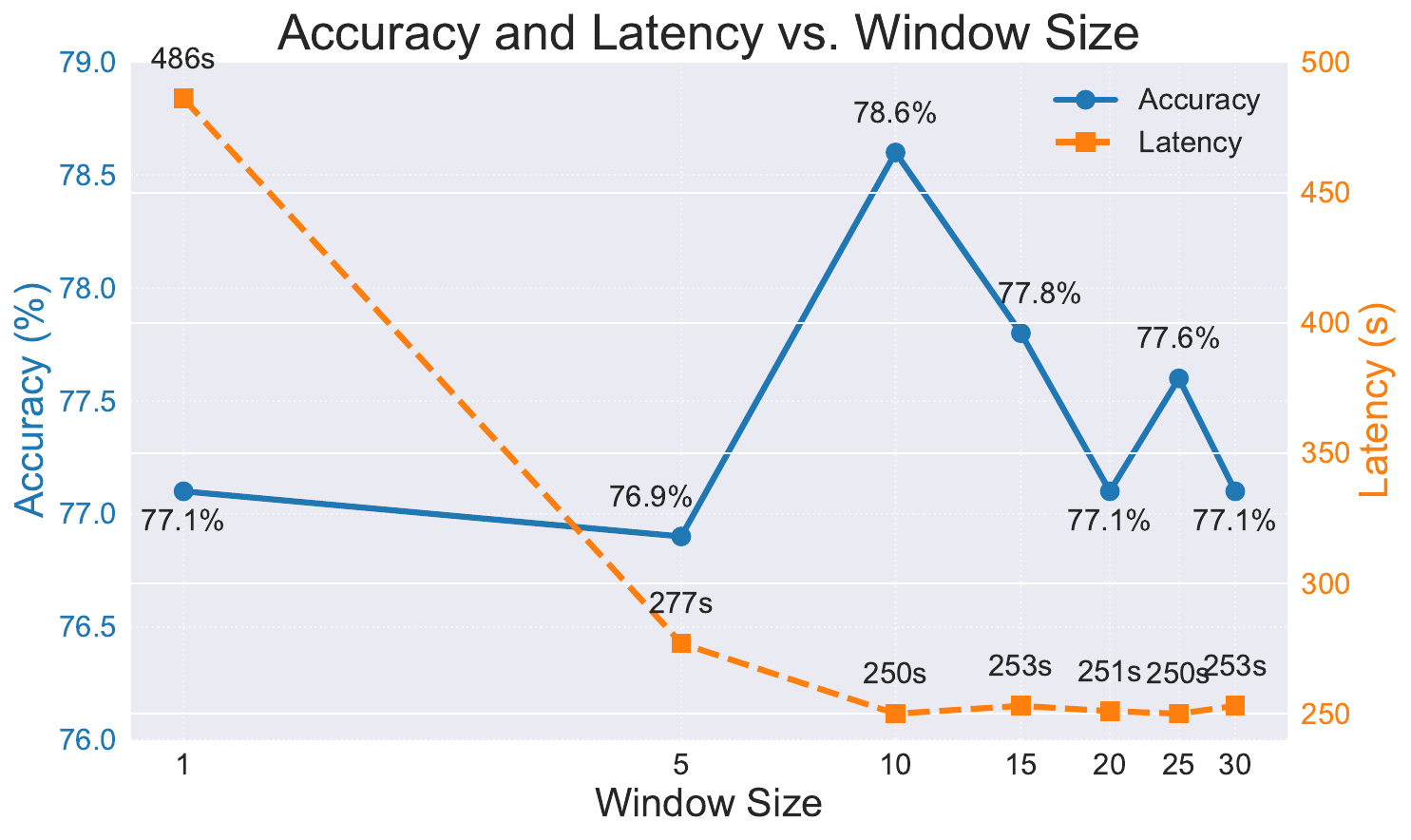}
        \captionof{figure}{\textbf{Accuracy and latency as functions of input window size:} The graph illustrates the interplay between model accuracy, processing latency, and the window size. Notably, accuracy peaks at a window size of 10, while latency stabilizes for window sizes of 10 and above. In all cases the accuracy only slightly fluctuates.}
        \label{fig:supp_window}
    \end{minipage}  
\end{table*}

\subsubsection{How does the number of frames affect the model's accuracy and latency?}
MovieChat \cite{song2024moviechat} processes 2048 frames for each video, while we use only 100 frames, as previous studies have demonstrated how redundant video data is \cite{simonyan2014two, wang2016temporal}. Given that the MovieChat-1k dataset contains very long videos (some exceeding 14,000 frames), we conducted experiments to extend the number of frames our model processes. Specifically, we experiment with 40, 80, 100, 300, 500, and 1000 frames while keeping the number of episodes constant. As for the \SubmoduleII\ keep ratio, we decrease it in function of the number frames so that the number of semantic features we keep equals 20.

We observe a complex relationship between model accuracy, processing latency, and the number of frames analyzed. Figure \ref{fig:acc_latency_frames} illustrates these relationships, providing insights into the performance trade-offs of our model. As evident from Figure \ref{fig:acc_latency_frames}, the relationship between accuracy and the number of frames is non-monotonic. Accuracy initially increases as the number of frames grows, reaching a peak of 79.4\% at 80 frames with a modest latency (note that we use 100 frames as the default parameter in other experiments for consistency with other datasets). This suggests that up to this point, additional frames provide valuable context that enhances the model's understanding. However, beyond 80 frames, we observe a decline in accuracy, possibly due to the introduction of noise or irrelevant information from temporally distant parts of the video.

Latency, on the other hand, exhibits a near-linear increase with the number of frames up to 300 frames, after which it grows super-linearly. This rapid increase in latency for higher frame counts underscores the computational challenges of processing large numbers of frames, particularly in real-time or near-real-time applications.

Interestingly, the model's performance at 1000 frames (76.7\% accuracy) is lower than its performance at 40 frames (77.6\% accuracy), but with a significantly higher latency (2676s vs. 143s). This observation highlights the diminishing returns and potential drawbacks of simply increasing the number of processed frames. It also underscores the importance of thoughtful frame selection in video understanding tasks. Future work could explore adaptive frame selection techniques that dynamically adjust the number of frames based on video content, potentially optimizing both accuracy and efficiency.

\subsubsection{How does the window size affect the model's accuracy and latency?}
Our analysis of our model's zero-shot performance on the MovieChat-1k test set reveals intriguing relationships between accuracy, latency, and input window size. Figure \ref{fig:supp_window} illustrates these trade-offs. As evident from Figure \ref{fig:supp_window}, the relationship between accuracy and window size is non-monotonic. Accuracy initially increases with window size, reaching a peak of 78.6\% at a window size of 10. This suggests that providing more context to the model improves its performance up to a certain point. However, beyond this optimal window size, accuracy begins to decline, possibly due to the introduction of irrelevant context.

Latency exhibits a sharp decrease from window size 1 to 5, after which it remains relatively stable. This indicates that while smaller window sizes may seem computationally advantageous, they incur higher latency, possibly due to the need for more frequent \SubmoduleI\ call. The optimal trade-off occurs at a window size of 10, where we observe peak accuracy and stabilized latency suggesting that carefully tuned context windows can enhance long-form video understanding without incurring additional computational costs.


\subsection{\ModelName~vs. MA-LMM vs. MovieChat}
\label{supp:vs}
\textbf{HERMES versus MA-LMM}: For each incoming frame, MA-LMM adds it to the memory bank by computing the similarities with adjacent frames and merging the incoming frame with its most similar in the memory bank. Below are our main differences.

\begin{itemize}
    \item HERMES takes a distributed approach. Our ECO, distributes the frames of the incoming window to the most appropriate episode. This approach is more intuitive and better mirrors human memory formation.
    \item Frames can be grouped into episodes regardless of temporal adjacency, unlike MA-LMM which only considers adjacent frames. This naturally handles scene transitions, flashbacks, and non-linear narratives.
    \item HERMES is vastly more efficient and accurate. As shown in \Cref{tab:submodules-malmm} in the main paper, our memory management system almost halves the inference time (-43\%) when plugged into MA-LMM while being 3.4\% more accurate.
    \item \ModelName~ also captures semantics. Our Semantics Retriever (SeTR) complements the episodic memory and is shown in \Cref{tab:submodules-malmm} to increase the accuracy of MA-LMM by almost 4\% with only a negligible increase in latency.
\end{itemize}

\noindent\textbf{HERMES versus MovieChat}: Moviechat's short-term memory uses a FIFO mechanism. Its long-term memory uses ToMe. Below are the main differences
\begin{itemize}
    \item HERMES has episodes instead of short-term memory, and our update approach is based on similarity to a certain existing episode instead of FIFO. As shown in \Cref{tab:\SubmoduleI} of the paper, FIFO's performance is inferior to ECO.
    \item \ModelName's long-term memory is implicitly encoded in ECO. We consider SeTR as a semantics scanner that retrieves scattered semantics from the video.
    \item 22 FPS processing speed compared to MovieChat's ~0.01 FPS (13 minutes vs 1 day on MovieChat-1k) using a V100 GPU (32 GB).
    \item \ModelName~ achieves high performance with only 100 frames compared to MovieChat's 2048 frames.
\end{itemize}

\subsection{A Note on Latency}
\label{appendix:latency}
The MovieChat-1k test set comprises 170 videos, from each of which our model samples 100 frames. This results in a total of 17,000 frames to be processed. Our empirical measurements show that the model requires 774 seconds to complete end-to-end inference on this dataset using a single V100 GPUs (32GB VRAM). This translates to a processing speed of approximately \textbf{22 frames per second (FPS)}, which is very close to real-time performance. Such a result suggests that our approach is not only effective in terms of accuracy but also efficient enough for practical applications in video understanding tasks.

\begin{figure*}[ht!]
    \centering
    \begin{subfigure}[b]{\textwidth}
        \centering
        \includegraphics[width=\textwidth]{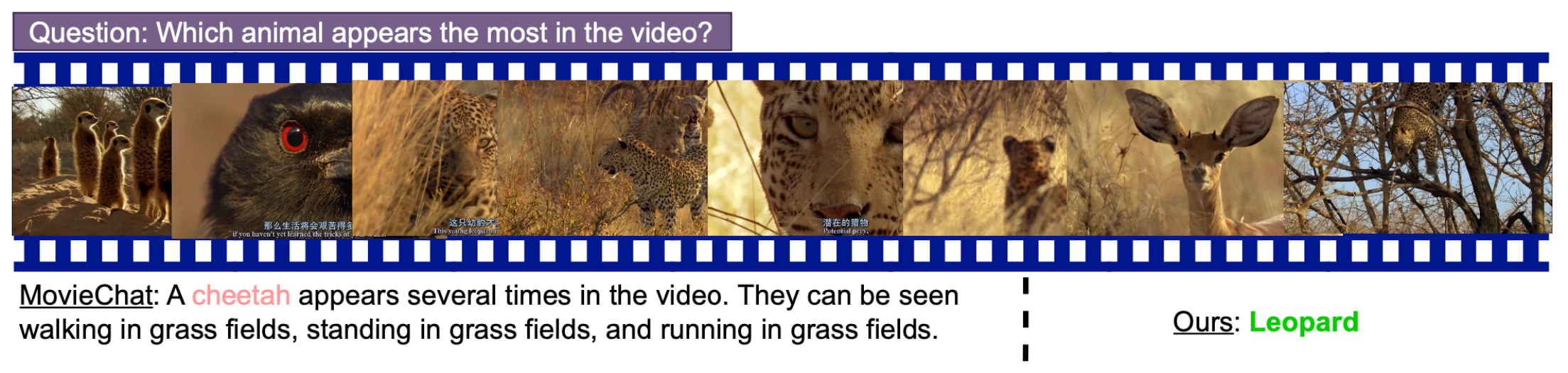}
        \caption{\textbf{Animal Identification:} MovieChat mistakenly identifies a Leopard as a Cheetah, even though no Cheetah appears in the video.}
        \label{fig:supp_qualitative_1}
    \end{subfigure}

    \begin{subfigure}[b]{\textwidth}
        \centering
        \includegraphics[width=\textwidth]{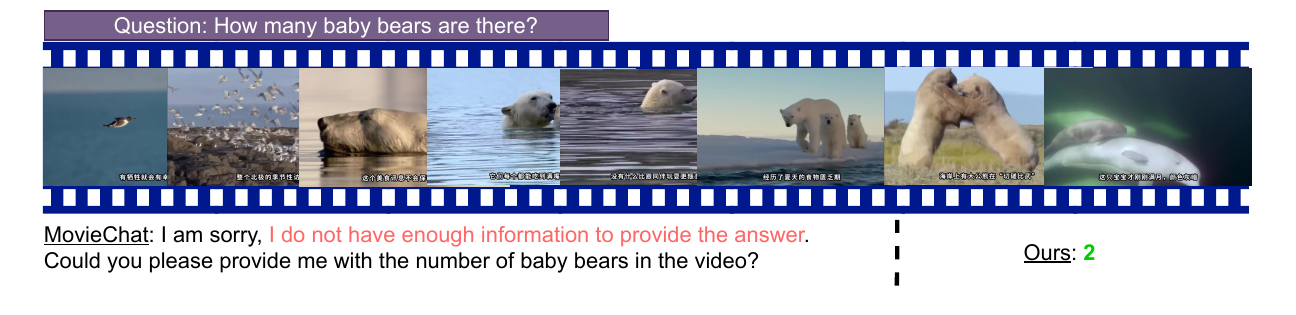}
        \caption{\textbf{Animal Counting:} This question is particularly challenging because the bears appear infrequently in the video, and the question specifically asks about ``baby bears." Despite MovieChat analyzing 2048 frames and our model only analyzing 100 frames, our model was able to locate and count the baby bears accurately.}

        \label{fig:supp_qualitative_2}
    \end{subfigure}

    \begin{subfigure}[b]{\textwidth}
        \centering
        \includegraphics[width=0.95\textwidth]{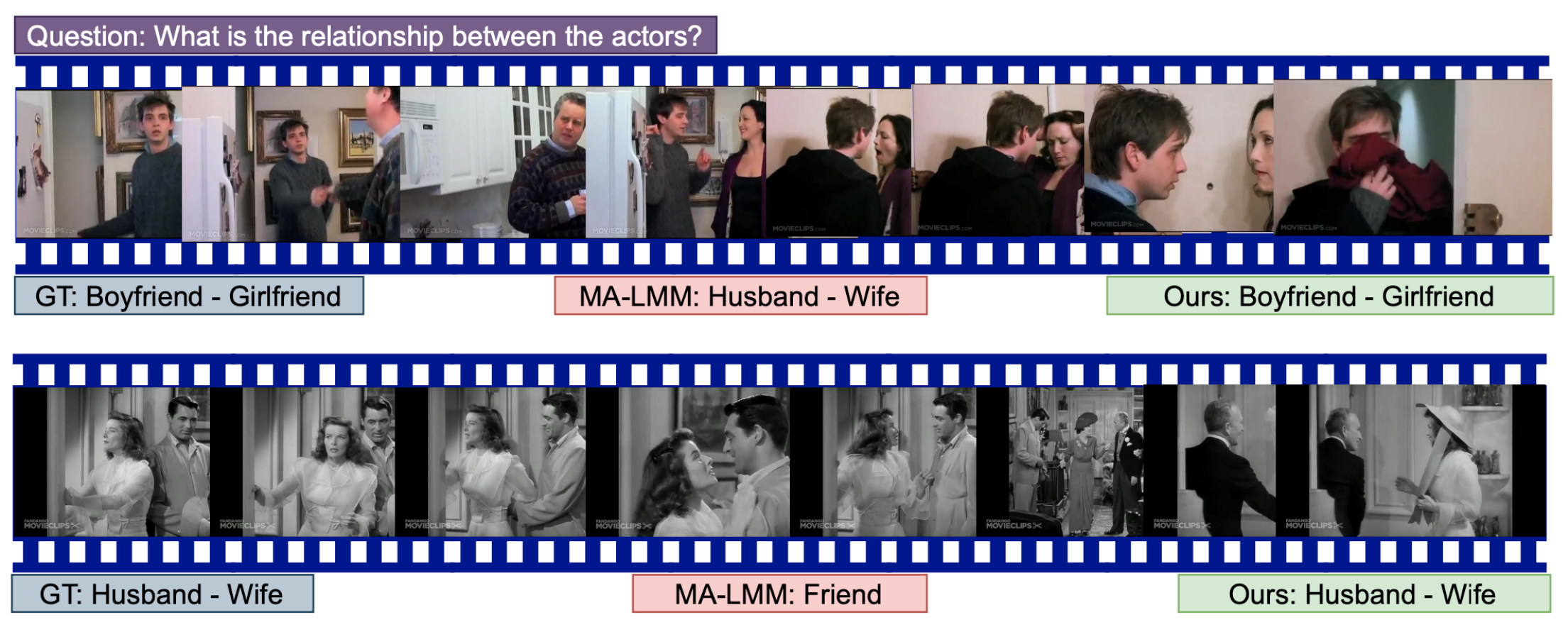}
        \caption{\textbf{Determining People's Relationships:} We compare our results with those of MA-LMM, with both models trained on the LVU dataset. Thanks to our episodic memory compression, our model excels at determining people's relationships across thousands of frames of interactions.}

        \label{fig:supp_qualitative_3}
    \end{subfigure}

    \caption{Qualitative results demonstrating the capabilities of our model compared to MovieChat and MA-LMM across different tasks. (a) Animal identification shows MovieChat's confusion between Leopard and Cheetah. (b) Animal counting highlights the challenge of locating baby bears with limited appearances in the video, where our model outperforms despite fewer frames. (c) Relationship determination benefits from our episodic memory compression, enabling better identification of relationships over extended interactions.}
    \label{fig:sup_qualitative}
\end{figure*}

\subsection{Qualitative Results}
\label{appendix:qual}

\noindent\textbf{Animal Identification.} Figure \ref{fig:supp_qualitative_1} demonstrates our model's superior performance in animal identification compared to MovieChat. In this example, MovieChat incorrectly identifies a leopard as a cheetah, despite no cheetah being present in the video. This misidentification underscores the importance of accurate visual feature extraction and semantic understanding in long-form video analysis.

\noindent\textbf{Animal Counting.} Figure \ref{fig:supp_qualitative_2} showcases our model's ability to perform complex counting tasks, even with limited information. The task involves counting baby bears, which appear infrequently in the video. Despite analyzing only 100 frames compared to MovieChat's 2048 frames, our model accurately locates and counts the baby bears. This demonstrates the efficiency of our \SubmoduleI\ and \SubmoduleII\ modules in capturing and retaining crucial information from sparse appearances.

\noindent\textbf{Determining People's Relationships.} In Figure \ref{fig:supp_qualitative_3}, we compare our model's performance against MA-LMM in determining relationships between people over extended video sequences. Both models were trained on the LVU dataset. Our model's superior performance in this task can be attributed to the episodic memory compression technique, which allows for better retention and analysis of interactions across thousands of frames.

\subsubsection{Visualization of \SubmoduleI{} and \SubmoduleII}
\label{supp:ecoviz}
\begin{figure*}[ht!]
    \centering
    \includegraphics[width=0.95\textwidth]{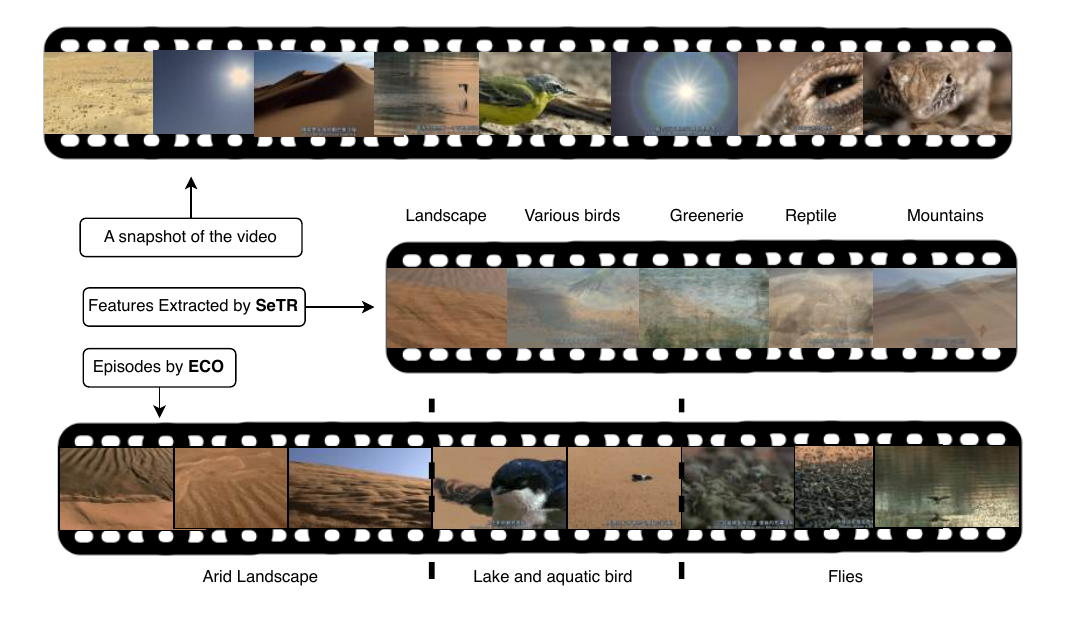}
    \caption{\textbf{Visualization of \SubmoduleI{} and \SubmoduleII:} The top row presents a curated visual summary of the video, showcasing key scenes such as landscapes, wildlife, and environmental features. The middle row highlights the functionality of \SubmoduleII{}, which extracts semantic features and clusters frames into thematic groups, including ``Landscape," ``Various Birds," and ``Reptiles." Finally, the bottom row illustrates the operation of \SubmoduleI{}, which segments the video into coherent narrative episodes, such as ``Arid Landscape," ``Lake and Aquatic Bird," and ``Flies." Together, these modules provide both high-level abstraction and detailed episodic structure for comprehensive video understanding.}
    \label{fig:supp_eco_viz}
\end{figure*}

Figure~\ref{fig:supp_eco_viz} demonstrates the inner-workings of \SubmoduleI{} and \SubmoduleII{}. The top row illustrates a curated summary of the video content, highlighting diverse scenes, such as landscapes, wildlife, and environmental features. 

\SubmoduleII{} is responsible for extracting high-level semantic features and grouping frames with similar themes, as shown in the mid row. For instance, the module effectively captures thematic clusters such as ``Landscape," ``Various Birds," and ``Reptiles," providing a concise overview of the video.

Meanwhile, \SubmoduleI{} processes the video at a more granular level, segmenting it into coherent episodes that reflect the narrative flow. The bottom row showcases this segmentation, organizing the content into episodic units like ``Arid Landscape," ``Lake and Aquatic Bird," and ``Flies." This two-tiered approach ensures both thematic abstraction and temporal coherence, enabling a comprehensive understanding of the video.

\subsection{Error Analysis: When does \ModelName\ fail and why?}
\label{sub:error}
\begin{figure*}[ht!]
    \centering
    \includegraphics[width=0.95\textwidth]{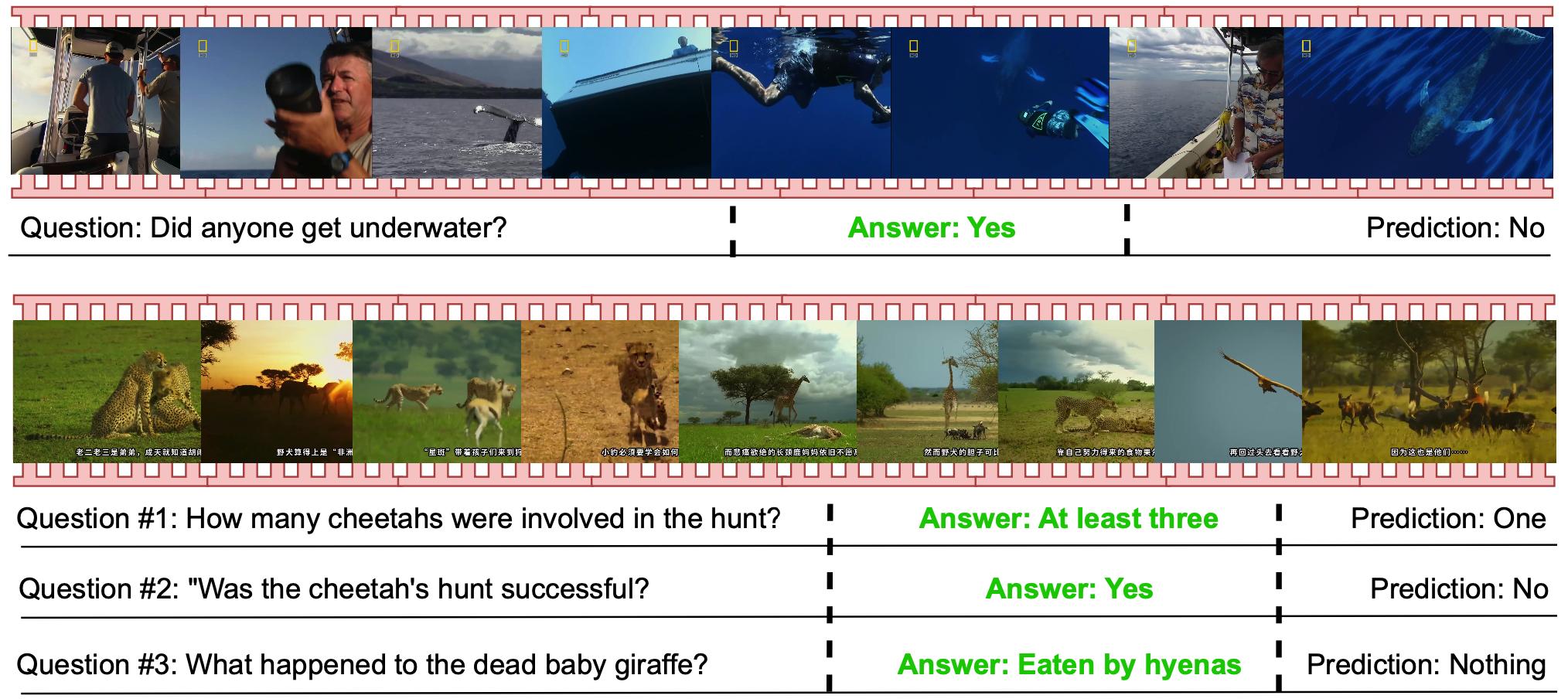}
    \caption{\textbf{Where and when \ModelName\ fail:} The top row shows a marine life video where the model fails to recognize underwater scenes. The bottom row depicts a wildlife documentary where the model struggles with quantitative reasoning and event inference across multiple frames. These cases highlight limitations in contextual understanding and temporal information integration.}
    \label{fig:supp_failure}
\end{figure*}

Our model, while generally effective, demonstrates several notable failure cases that warrant further investigation and improvement. Figure \ref{fig:supp_failure} illustrates examples where the model's predictions deviate from ground truth answers, revealing key limitations in contextual reasoning and temporal information integration. Figure \ref{fig:supp_failure} presents two sets of video frame sequences that highlight shortcomings in our model's performance. In the top row, we observe a documentary on marine life. Despite clear visual cues of underwater scenes and diving equipment, the model incorrectly predicts that no one got underwater. The bottom row showcases a more complex scenario from a wildlife documentary. Here, the model exhibits multiple errors: It underestimates the number of cheetahs involved in the hunt, predicting only one when at least three are present. This indicates a weakness in quantitative reasoning across temporally distributed information. The model incorrectly predicts that the cheetah's hunt was unsuccessful, contradicting the visual evidence. This error points to difficulties in inferring outcomes from sequences of events. Lastly, the model fails to recognize the fate of a dead baby giraffe, predicting ``nothing" when the correct answer is ``eaten by hyenas". 

These examples emphasize the need for improved mechanisms to aggregate and reason over long-range temporal dependencies, as well as enhanced capabilities in scene understanding and event inference. 

\subsection{How is our approach related to cognitive processes?}
\label{cognition}
Our approach to long-form video understanding is inspired by cognitive processes involving memory and comprehension. According to the literature on neuroscience \cite{tulving1972episodic, schacter1982memory, tulving1983elements}, human cognition involves two primary types of memory: episodic and semantic. Episodic memory is the ability to recall specific events or episodes, while semantic memory refers to the storage of general knowledge and concepts. These forms of memory are crucial for understanding long-form narratives, where a coherent understanding arises from the integration of specific events and overarching themes.

The proposed \ModelName\ model incorporates these cognitive processes through its two main components, \SubmoduleI\ and \SubmoduleII. \SubmoduleI, akin to the function of episodic memory, selectively retains and compresses key events from the video, allowing the model to form a structured representation of the narrative as it unfolds. This approach is an oversimplified abstraction of findings in cognitive neuroscience, which highlight the role of the hippocampus in the consolidation of episodic memories \cite{eichenbaum2004hippocampus, schacter1982memory}, and the concept of \textit{subjective time} \cite{arstila2014subjective} that sees a scene (or a video) not as a series of frames but as a series of experiences. The hippocampus enables the organization of temporally distinct experiences into a coherent memory trace, something that we aim to capture with \SubmoduleI. Moreover, the sequential processing and aggregation of information in our model align with the concept of event segmentation in cognitive psychology \cite{zacks2007event}. Humans naturally segment continuous experiences into discrete events, which aids in memory formation and recall.

Meanwhile, \SubmoduleII\ functions similarly to semantic memory, extracting and reinforcing high-level semantic cues. This process mirrors how the brain integrates detailed episodic memories with broader semantic knowledge stored in the neocortex \cite{mcclelland1995there, binder2011neurobiology}. Also related is the concept of gist extraction which involves rapidly comprehending the essence or overall meaning of a scene or situation \cite{oliva2005gist}. This ability allows humans to quickly understand the context of a complex scene without processing every detail. Our \SubmoduleII\ operates similarly by identifying and extracting high-level semantic cues that provide a concise overview of the scene and actions.

The integration of these cognitive processes not only aligns with human-like comprehension but also offers a framework for efficiently handling the vast and diverse information present in long-form videos. Significant improvements over existing state-of-the-art models, underscore the effectiveness of this cognition-inspired approach. While our model is a oversimplified abstraction of human cognition, it provides a foundation for exploring more complex cognitive mechanisms in future work.

\end{document}